\documentclass{eptcs}
\providecommand{\event}{TARK 2017} 
\usepackage{underscore}           

\usepackage{amsmath,tikz}

\title{Cheryl's Birthday}
\author{Hans van Ditmarsch
\institute{LORIA, CNRS\\
University of Lorraine\\
Nancy, France}
\email{hans.van-ditmarsch@loria.fr}
\and
Michael Ian Hartley
\institute{DownUnder Geosolutions\\
Sydney, Australia}
\email{mikeh@dug.com}
\and
Barteld Kooi
\institute{Faculty of Philosophy\\
University of Groningen\\
 Netherlands}
\email{b.p.kooi@rug.nl}
\and
Jonathan Welton
\institute{95 Heathermount Drive \\ Crowthorne, Berkshire \\ United Kingdom} 
\email{j\_welton@hotmail.com}
\and
Joseph B.W.\ Yeo
\institute{National Institute of Education \\ Nanyang Technological University \\
Singapore} 
\email{josephbw.yeo@nie.edu.sg}
}

\def\titlerunning{Cheryl's Birthday}
\def\authorrunning{H.\ van Ditmarsch, M.I.\ Hartley, B.\ Kooi, J.
  Welton \& J.B.W. Yeo}
\begin{document}
\maketitle

\begin{abstract}
We present four logic puzzles and after that their solutions. Joseph Yeo designed {\em Cheryl's Birthday}. Mike Hartley came up with a novel solution for {\em One Hundred Prisoners and a Light Bulb}. Jonathan Welton designed {\em A Blind Guess} and {\em Abby's Birthday}. Hans van Ditmarsch and Barteld Kooi authored the puzzlebook {\em One Hundred Prisoners and a Light Bulb} \cite{hvdetal.puzzle:2015} that contains other knowledge puzzles, and that can also be found on the webpage \url{http://personal.us.es/hvd/lightbulb.html} dedicated to the book. 
\end{abstract}

\section{Logic Puzzles}

\subsection{Cheryl's Birthday}

{\em 
Albert and Bernard just became friends with Cheryl, and they want to know when her birthday is. Cheryl gives them a list of 10 possible dates for her birthday: May 15, 16, 19; June 17, 18; July 14, 16; and August 14, 15, 17. Cheryl then tells only to Albert the month of her birthday, and tells only to Bernard the day of her birthday. (And Albert and Bernard are aware that she did so.) Albert and Bernard now have the following conversation:
\begin{itemize}
\item Albert: ``I don't know when Cheryl's birthday is, but I know that you don't know either.''
\item Bernard: ``At first I didn't know when Cheryl's birthday is, but now I know.''
\item Albert: ``Now I also know when Cheryl's birthday is.''
\end{itemize}
{\bf When is Cheryl's birthday?}
}

\medskip

This riddle was set by Joseph Yeo as one of the problems for the Secondary $3$ and the Secondary $4$ levels of the $2015$ Singapore and Asian Schools Math Olympiad (SASMO) contests. He had modified it from a 2013 entry of a blog by `Mr Brown',\footnote{\url{http://www.mrbrown.com/blog/2013/02/yet-another-psle-} \url{maths-question-to-blow-your-mind.html}} that Joseph later traced to a 2006 entry in {\em The Math Forum `Ask Dr.\ Math'}.\footnote{\url{http://mathforum.org/library/drmath/view/68613.html}} Cheryl's Birthday riddle is reminiscent of Freudenthal's Sum and Product riddle \cite{freudenthal:1969} (see  also \cite[Chapter 7]{hvdetal.puzzle:2015}), wherein the conversation between Mr.\ Sum and Mr.\ Product is very similar.

Cheryl's Birthday riddle subsequently caught a lot of public attention. A Singapore TV presenter, Kenneth Kong, was given this puzzle by a friend whose niece had difficulty solving it. Because the niece was in primary school, Kong thought that the riddle was intended for primary school students. And because Kong and his wife disagreed on the correct answer, he decided to post it online on his Facebook page. This happened on 10 April 2015. Then, as they say, it went viral. It was in subsequent days discussed by Alex Bellos in the British newspaper The Guardian, in the BBC news, in the New York Times, in India Today, etc.\footnote{\url{https://en.wikipedia.org/wiki/Cheryl's_Birthday}} See also \cite{yeo:2016}.

\subsection{One Hundred Prisoners and a Light Bulb}

{\em A group of 100 prisoners, all together in the prison dining area, are told that they will be all put in isolation cells and then will be interrogated one by one in a room containing a light with an on/off switch. The prisoners may communicate with one another by toggling the light-switch (and that is the only way in which they can communicate). The light is initially switched off. There is no fixed order of interrogation, or interval between interrogations, and the same prisoner may be interrogated again at any stage. When interrogated, a prisoner can either do nothing, or toggle the light-switch, or announce that all prisoners have been interrogated. If that announcement is true, the prisoners will (all) be set free, but if it is false, they will all be executed. While still in the dining room, and before the prisoners go to their isolation cells (forever), can the prisoners agree on a protocol that will set them free?}

\medskip

The 100 prisoners riddle has been going round since the early 2000s, a very nice early elaboration is \cite{dehayeetal:2004}. Solutions wherein agents play different roles (`counter' and `follower') are widely known, and also solutions wherein all agents play the same role but wherein part of their behaviour is determined by chance (whether one becomes a `counter' or a `follower' is determined by rolling a pair of dice), and wherein termination is not guaranteed but expected (the probability will approach 1). When Hans van Ditmarsch gave a public interest lecture on {\em One Hundred Prisoners and a Light Bulb} at the University of Western Australia, in Perth, Mike Hartley --- he also runs a math puzzle site for children\footnote{\url{http://www.dr-mikes-math-games-for-kids.com/}} --- proposed a solution wherein all agents play the same role (behave the same) and where their behaviour is still determined and not probabilistic. Also in that case, termination is also not guaranteed but expected. 

\begin{quote}
{\bf What uniform protocol can they follow?} 
\end{quote}

\subsection{A Blind Guess}

{\em 
There are three players, each with a card bearing a single digit ($0-9$) on their head such that they can see the other two numbers but not their own --- although one is blind and can see no cards. They are told the three digits make up a three digit square number (i.e. between $10 \cdot 10$ and $32 \cdot 32$), and are asked to call out when they know their own number. After a long, long, long silence, the blind person correctly identifies his card's number. 

{\bf What was it?}
}

\medskip

This riddle was proposed by Jonathan Welton. It can be seen as a variation of the {\em Who has the Sum?} riddle \cite[Chapter 6]{hvdetal.puzzle:2015}. That riddle was also originally proposed by him, in the {\em Sunday Times} magazine Brainteaser column, in the mid 1990s, and was later reprinted as Puzzle 27 in {\em The Sunday Times Brainteasers} by Victor Bryant \cite{bryant:2002}. Hans and Barteld did not know this when they wrote their puzzlebook, although they made an effort to trace the origin of the riddles and to report on their earliest known source (their source for this riddle was from 2003). But they found out when Jonathan contacted them, reading about it in the book.

\subsection{Abby's Birthday}

{\em 
The problem setter announces to two strangers Abby and Barry: \begin{quote} ``You each know the day of the week you were born on, but not the day of the week each other was born on. However, I can tell you that you were born a day apart, and that Barry was not born on a Monday.'' \end{quote} He then proceeds to ask Abby and Barry alternately if they can deduce which day of the week the other person was born on, and receives the answers:
\begin{itemize}
\item Abby: ``No.''  
\item Barry: ``No.''  
\item Abby: ``No.''  
\item Barry: ``No.'' 
\item Abby: ``No.''
\end{itemize}
{\bf On what day of the week was Abby born?}
}

\medskip

Abby's Birthday problem was also proposed by Jonathan Welton.

\section{Solutions}

We now proceed with the solutions of the riddles. It is more fun if you first try to solve them yourself.

\subsection{Cheryl's Birthday}

We can visually represent the situation wherein
Albert and Bernard have just been told their private information about
the month and the day, respectively, as follows.

\medskip

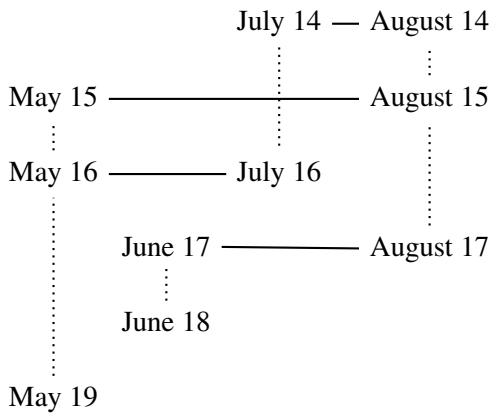
\begin{tikzpicture}[thick]
\node[anchor=mid] (may19th) at (0,0) {May 19};
\node[anchor=mid] (may16th) at (0,3) {May 16};
\node[anchor=mid] (may15th) at (0,4) {May 15};
\node[anchor=mid] (june18th) at (1.5,1) {June 18};
\node[anchor=mid] (june17th) at (1.5,2) {June 17};
\node[anchor=mid] (july14th) at (3,5) {July 14};
\node[anchor=mid] (july16th) at (3,3) {July 16};
\node[anchor=mid] (august17th) at (5,2) {August 17};
\node[anchor=mid] (august15th) at (5,4) {August 15};
\node[anchor=mid] (august14th) at (5,5) {August 14};

\draw[dotted] (may19th) edge (may16th);
\draw[dotted] (may16th) edge (may15th);
\draw[dotted] (june18th) edge (june17th);
\draw[dotted] (july14th) edge (july16th);
\draw[dotted] (august17th) edge (august15th);
\draw[dotted] (august15th) edge (august14th);

\draw (july14th) edge (august14th);
\draw (may15th) edge (august15th);
\draw (may16th) edge (july16th);
\draw (june17th) edge (august17th);
\end{tikzpicture}

\medskip

Dates that share the same month are connected by dotted
lines. These indicate that Albert cannot distinguish these. When two
dates have the same day, they are connected by a solid line,
indicating that Bernard cannot distinguish them.

When you look at this picture, June 18th and May 19th are special in
the sense that they are not connected to other dates by solid lines.
That means that if one of those dates were Cheryl's birthday, then
Bernard would know her birthday, because there are no alternatives
for him to consider.

Note also that in this picture for any date there is another date connected to it by a
dotted line. This means that for Albert, there is always an
alternative. So, whatever Cheryl's birthday is, Albert won't know
when. This also means that when he says that he does not know her
birthday, he does not convey any information. This was common knowledge
even to an outsider. However, when Albert says that he knows that
Bernard does not know Cheryl's birthday, we can infer that Cheryl's birthday is not in
May or June, because if it were in May or June --- as far as Albert is
concerned --- her birthday might have been on the 19th or 18th and then
Bernard would know. The only way that Albert can know that Bernard does not
know the birthday, is if this is not the case. 

If we update the
model with the information that Cheryl's birthday is not in May or June we find the following picture.

\medskip

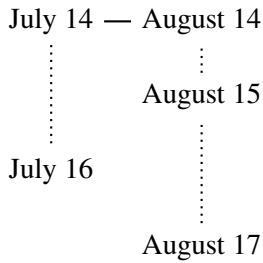
\begin{tikzpicture}[thick]
\node[anchor=mid] (july14th) at (3,5) {July 14};
\node[anchor=mid] (july16th) at (3,3) {July 16};
\node[anchor=mid] (august17th) at (5,2) {August 17};
\node[anchor=mid] (august15th) at (5,4) {August 15};
\node[anchor=mid] (august14th) at (5,5) {August 14};

\draw[dotted] (july14th) edge (july16th);
\draw[dotted] (august17th) edge (august15th);
\draw[dotted] (august15th) edge (august14th);

\draw (july14th) edge (august14th);
\end{tikzpicture}

\medskip

In this picture the only dates connected by a solid lines are July
14th and August 14. Consequently, on those dates Bernard does not know
when Cheryl's birthday is, but on all of the other dates he would
know, since in the updated picture there are no alternative dates for
him to consider. (Not only did we learn from Albert's assertion that
May and June are excluded, Bernard learnt the same.) So when Bernard
states that he does now know when Cheryl's birthday is, we can
conclude her birthday is not on the 14th. (We can
disregard his statement that at first he did not know Cheryl's birthday, because Albert already asserted that.) 

If we carry out the update we get the following picture.

\medskip

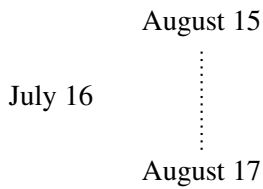
\begin{tikzpicture}[thick]
\node[anchor=mid] (july16th) at (3,3) {July 16};
\node[anchor=mid] (august17th) at (5,2) {August 17};
\node[anchor=mid] (august15th) at (5,4) {August 15};

\draw[dotted] (august17th) edge (august15th);

\end{tikzpicture}

\medskip

In this picture we see that Albert is still not certain of Cheryl's
birthday if it is in August, but otherwise he knows her birthday,
because there are no alternatives. He states that he knows, and
therefore we now also know that Cheryl's birthday is on:

\medskip

July 16

\medskip

Joseph Yeo ventures that {\em Cheryl's Birthday} was so widely discussed in the media because there was much discussion whether the answer should be July 16 or August 17. In \cite[page 4]{yeo:2016} he explains in detail how some people obtained the wrong answer August 17.

\subsection{One Hundred Prisoners and a Light Bulb}

A well-known solution consists of the prisoners appointing one of them as the {\em counter}, where the remaining prisoners are {\em followers}. A follower turns the light on when it is off, but does so only once. If the light is on he does nothing and if the light is off and he has turned it on before, he also does nothing. The counter turns the light off when it is on, and keeps count of how many times he has turned the light off; and does nothing when the light is on. When the counter has reached 100 he announces that everybody has been interrogated. In this standard solution we can think of all the prisoners as initially having a token with value 1 in their head. The followers will drop one of their token when they turn on the light and remain with value 0. Whereas the counter, by turning off the light, only increases his token all the time, until it has value 100.

There is a probabilistic solution where all prisoners can both act as the counter and as a follower but do so depending on the value of their token and depending on the outcome of rolling a dice: given the value of their token and the state of the light they may {\em either}, \begin{quote} when the light is off, turn on the light and reduce their token, and, when the light is on, do nothing (thus acting as a follower); \end{quote} {\em or}, \begin{quote} when the light is off, do nothing and when the light is on, turn the light off and increase their token (thus acting as a counter).\end{quote} Termination cannot be guaranteed but can be expected (the probability will approach 1 in due time). Expected execution time (given random scheduling of interrogations) can be shortened if the probability of being a counter is high (or even 1) once a prisoner holds a token with value more than 50 (although this has not been tested by simulations or verified by calculations). Roles are not fixed: for token values below $50$ it is important that the probability of being counter or follower is not 0 and not 1. For example, if there are two prisoners hold tokens of value 50, and they only act as counters for that value of token, neither of them will ever turn on the light. Nobody else will anyway. So then the interrogations will go on forever without termination. No prisoner will reach token value 100. For more details on this solution, see \cite[page 92, Protocol 6]{hvdetal.puzzle:2015}.

It came therefore as a surprise to Hans that Mike Hartley suggested a solution wherein all prisoners play the same role (and wherein that role is not probabilistic), i.e., wherein the state of the light and the value of their token determine whether they act as counter or as follower. It is as follows. 

\begin{itemize}
\item Each prisoner initially holds a token valued 1.
\item If a prisoner enters the room and the light is off,
\begin{itemize}
\item then if their token is 0, they do nothing;
\item otherwise, they switch the light on and subtract 1 from their token.
\end{itemize}
\item If a prisoner enters the room and the light is on,
\begin{itemize}
\item then if their token is 0, they do nothing;
\item and if their token is 99, they announce that all prisoners have been interrogated;
\item otherwise, they switch off the light and add 1 to their token.
\end{itemize}
\end{itemize}
Given random scheduling this protocol is again expected to terminate. This may be somewhat surprising, as `on average' a prisoner will turn the light off equally often as turning the light on. The reason is the (known) deviation from this expected average. A large deviation may result in a prisoner substantially increasing his token, for a while, after which it may just as well substantially decrease again. Once this deviation hits the roof, token value 100, the protocol terminates. The longer the interrogations, the more likely it becomes to hit the roof. Clearly, this may take a while. 

Now for the standard protocol, given uniform random scheduling, and on assumption a single interrogation takes place every day, the prisoners can expect to get out of prison after 24,5 years. For the prisoners, who of course only care to get out of prison as soon as possible, this outlook is already not quite so rosy. But this novel protocol is much worse. Mike Hartley:

\begin{quote} {\em 
I ran a computer simulation of it just now, and the $100$ prisoners got out in, on average, 1613 years. The luckiest group got out in `only' $137$ years, the unluckiest were imprisoned when the ancient Sumerians were just getting started, and still have $1800$ years of their sentences to serve.
}
\end{quote}

Suppose $k$ out of $n$ prisoners have token values $a_1$, $a_2$, \dots, $a_k$, the rest having token value 0. Let the expected time for the prisoners to escape be $U_{a_1, a_2, \dots a_k}$ if the light is off, and $T_{a_1, a_2, \dots a_k}$ if the light is on. Note that $U_n=0$, since in that case the prisoners have already escaped. If a random prisoner is chosen to be interrogated, then 
\[\begin{array}{lll}
U_{a_1, a_2, \dots a_k} & = & 1 + \frac{n-k}{n}U_{a_1, a_2, \dots a_k}
+ 
\frac{1}{n}\sum_{i=1}^{k}T_{a_1-\delta_{1i}, a_2-\delta_{2i}, \dots, a_k-\delta_{ki}} \\ \ \\
\hfill \text{and} \\ \ \\
T_{a_1, a_2, \dots a_k} &= & 1 + \frac{n-k}{n}T_{a_1, a_2, \dots a_k}
+ 
\frac{1}{n}\sum_{i=1}^{k}U_{a_1+\delta_{1i}, a_2+\delta_{2i}, \dots, a_k+\delta_{ki}}.
\end{array}\]
This gives a system of linear equations for the expected times. It is hard to find general formulae for $U_{a_1, a_2, \dots a_k}$ and $T_{a_1, a_2, \dots a_k}$, but it is clear that the solution is finite, and so the prisoners will not be imprisoned forever. 

For example, to show an endgame, consider we have reached in the $100$ prisoners case that one prisoner has a token worth $99$, and that the light is on. We now get that $T_{99} = 1 + \frac{99}{100} T_{99} + \frac{1}{100} U_{(99+1)}$. Given that $U_{100} = 0$, $T_{99}$ must therefore be $100$. What does this mean? Given a prisoner with token $99$, the light on (token $1$) and all other $99$ prisoners with token $0$, there is a $\frac{1}{100}$ chance that the token-$99$ prisoner is interrogated and collects the last token, and there is a $\frac{99}{100}$ chance that a token-$0$ prisoner is interrogated, in which case the light remains on (back to $T_{99}$). The probability of $\frac{1}{100}$ indeed comes with an expectation of $100$ days. The expected time to escape for $U_{100}$ is $0$ because the prisoner with token $100$ declares victory: all prisoners have been interrogated. 

For three prisoners we get that $T_{a,b,c}$ equals: \[\begin{array}{lll}
\left(1+\frac{1}{a+b+c}\right)\cdot \left[2(a^2b+b^2c+c^2a+ab^2+bc^2+ca^2)+3abc+a+b+c\right] \end{array}\] so that $T_{1,1,1} = \left(1+\frac{1}{3}\right)\left[12+3+3\right] = 24$. To have to wait $24$ days to get out of prison is not so bad.


\subsection{A Blind Guess}

There are $22$ square numbers between $100$ and $999$. Every agent carries a number that is one of the three digits of such a square number. A model of the uncertainty about the squares therefore consists triples of digits. Now these are again, in a way, numbers. However, they are numbers between $1$ and $999$ --- where $1$ is in fact $001$. So, to model this problem we need far more than $22$ states. Let the agents be $A,B,C$ and let $C$ be the blind agent. We only need to model the uncertainty of $A$ and $B$. Blind agent $C$ plays the role of us, the problem solvers. The domain is partitioned into a great deal of parts that are $\{A,B\}$-connected and that are all disconnected from each other. Let us consider some such parts. Agent $A$ cannot distinguish the states connected by a solid line and agent $B$ cannot distinguish the states connected by a dotted line.

\medskip

\begin{tikzpicture}[thick]
\node (1) at (0,0) {100};
\node (2) at (1.5,0) {400};
\node (3) at (3,0) {900};
\draw[-] (1) to (2);
\draw[-] (2) to (3);
\end{tikzpicture}

\medskip

Here, $A$ cannot distinguish three states wherein she sees that $B$ and $C$ both have $0$. She is uncertain whether she has $1$, $4$, or $9$. But $B$ can distinguish all states. So, when asked to call out if he knows his number, he will say: ``Yes.'' But the problem description tells us he did not. We can therefore rule out the states $100$, $400$ and $900$ from the model of uncertainty. We cannot {\em yet} rule out those squares. For that, we also need to consider the part of the model

\medskip

\begin{tikzpicture}[thick]
\node (1) at (0,0) {010};
\node (2) at (1.5,0) {040};
\node (3) at (3,0) {090};
\draw[dotted,-] (1) to (2);
\draw[dotted,-] (2) to (3);
\end{tikzpicture}

\medskip

\noindent in which case $A$ will say that she knows her number, and the three $\{A,B\}$-disconnected parts of the model
 
\medskip

\begin{tikzpicture}[thick]
\node (1) at (0,0) {001};
\node (2) at (1.5,0) {004};
\node (3) at (3,0) {009};
\end{tikzpicture}

\medskip

\noindent in which case both $A$ and $B$ will call out that they know their number.

More interesting cases are, for example,

\medskip

\begin{tikzpicture}[thick]
\node (1) at (0,0) {526};
\node (2) at (1.5,0) {576};
\node (3) at (3,0) {676};
\draw[dotted,-] (1) to (2);
\draw[-] (2) to (3);
\end{tikzpicture}

\medskip

\noindent and

\medskip

\begin{tikzpicture}[thick]
\node (1) at (0,0) {675};
\node (2) at (1.5,0) {625};
\node (3) at (3,0) {225};
\node (4) at (4.5,0) {265};
\node (5) at (6,0) {765};
\draw[dotted,-] (1) to (2);
\draw[-] (2) to (3);
\draw[dotted,-] (3) to (4);
\draw[-] (4) to (5);
\end{tikzpicture}

\medskip

Note that the squares $625$ and $576$ occur in these parts of the model in all their permutations. Still, the parts do not have any symmetry. 

If we take the first part, then either $A$ or $B$ would have initially called out (namely if the state had been, respectively, $526$  or $676$), or else both would have called out later. But as this did not happen, after two updates this part of the model disappears. 

Similarly for the other part, after three updates. Agent $C$ then reasons at follows: the state cannot be $675$ because then $A$ would have called out. But it cannot be $625$ either because then, after $B$ does not hear $A$ call out, $B$ would have called out. And it cannot be $225$ either, because, after $A$ does not hear $B$ call out because $B$ did not hear $A$ call out, $A$ would have called out. And dually for $B$, from the other side of this $\{A,B\}$-chain.

We did not depict various other parts of the domain. After a certain finite number of such updates, there is no further uncertainty reduction. The domain will then be reduced to the following, unique, circular chain of uncertainty. 

\medskip

\begin{tikzpicture}[thick]
\node (1) at (0,0) {484};
\node (2) at (1.5,0) {184};
\node (3) at (3,0) {144};
\node (4) at (3,-1) {844};
\node (5) at (1.5,-1) {814};
\node (6) at (0,-1) {414};
\draw[-] (1) to (2);
\draw[dotted,-] (2) to (3);
\draw[-] (3) to (4);
\draw[dotted,-] (4) to (5);
\draw[-] (5) to (6);
\draw[dotted,-] (6) to (1);
\end{tikzpicture}

\medskip

\noindent We observe that $C$ holds $4$ in all these states. This is therefore the solution of the problem. 

We could actually think of $C$ as having the universal relation on the domain --- he is unable to distinguish any of the states. Successive absence of announcements of knowledge will shrink the domain more and more until at some stage only the circular partition remains. In this we have taken some `mathematical licence' interpreting the problem formulation. There is no explicit synchronization in the `after a long, long silence' part of the description. (By the way, this is the case in all older formulations of such knowledge and ignorance puzzles, such as Littlewood's 1953 {\em Miscellany} \cite{littlewood:1953}, and  prior to that). We could alternatively have formulated it more explicitly: the problem setter asks $A$ and $B$ to give a shout when he claps his hands, and tells them that he will continue to repeat this clapping action until \dots Until nothing happens? Then you can wait a long time! We would then have to specify the number of repetitions of hand clapping, like $3$, or $4$ ($3$ is in fact enough). We find the original formulation then more elegant, as the `long, long silence' suggests a {\em finite} number of iterations, without specifying the amount. Which is all the information we need as problem solvers.

\subsection{Abby's Birthday}

We number the days of the week from 1 to 7 for Monday to Sunday. Let a pair $ij$ stand for  `Abby was born on a $i$ and Barry was born on a $j$'. The model of uncertainty then is as follows, where Abby is uncertain between worlds linked by a solid line and Barry is uncertain between states linked by a dotted line.

\medskip

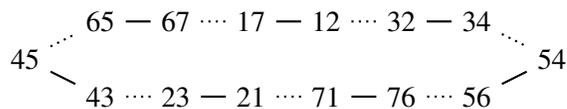
\begin{tikzpicture}[thick]
\node (17) at (0,0) {17};
\node (12) at (1,0) {12};
\node (32) at (2,0) {32};
\node (34) at (3,0) {34};
\node (54) at (4,-.5) {54};
\node (56) at (3,-1) {56};
\node (76) at (2,-1) {76};
\node (71) at (1,-1) {71};
\node (21) at (0,-1) {21};
\node (23) at (-1,-1) {23};
\node (43) at (-2,-1) {43};
\node (45) at (-3,-.5) {45};
\node (65) at (-2,0) {65};
\node (67) at (-1,0) {67};
\draw[-] (17) to (12);
\draw[dotted,-] (12) to (32);
\draw[-] (32) to (34);
\draw[dotted,-] (34) to (54);
\draw[-] (54) to (56);
\draw[dotted,-] (56) to (76);
\draw[-] (76) to (71);
\draw[dotted,-] (71) to (21);
\draw[-] (21) to (23);
\draw[dotted,-] (23) to (43);
\draw[-] (43) to (45);
\draw[dotted,-] (45) to (65);
\draw[-] (65) to (67);
\draw[dotted,-] (67) to (17);
\end{tikzpicture}

\medskip

The successive announcements result in the following uncertainty reductions.

\begin{itemize}
\item Problem setter: ``Barry was not born on Monday.''
\end{itemize}

\begin{tikzpicture}[thick]
\node (17) at (0,0) {17};
\node (12) at (1,0) {12};
\node (32) at (2,0) {32};
\node (34) at (3,0) {34};
\node (54) at (4,-.5) {54};
\node (56) at (3,-1) {56};
\node (76) at (2,-1) {76};
\node (23) at (-1,-1) {23};
\node (43) at (-2,-1) {43};
\node (45) at (-3,-.5) {45};
\node (65) at (-2,0) {65};
\node (67) at (-1,0) {67};
\draw[-] (17) to (12);
\draw[dotted,-] (12) to (32);
\draw[-] (32) to (34);
\draw[dotted,-] (34) to (54);
\draw[-] (54) to (56);
\draw[dotted,-] (56) to (76);
\draw[dotted,-] (23) to (43);
\draw[-] (43) to (45);
\draw[dotted,-] (45) to (65);
\draw[-] (65) to (67);
\draw[dotted,-] (67) to (17);
\end{tikzpicture}

\begin{itemize}
\item Abby: ``No.''  
\end{itemize}

\begin{tikzpicture}[thick]
\node (17) at (0,0) {17};
\node (12) at (1,0) {12};
\node (32) at (2,0) {32};
\node (34) at (3,0) {34};
\node (54) at (4,-.5) {54};
\node (56) at (3,-1) {56};
\node (43) at (-2,-1) {43};
\node (45) at (-3,-.5) {45};
\node (65) at (-2,0) {65};
\node (67) at (-1,0) {67};
\draw[-] (17) to (12);
\draw[dotted,-] (12) to (32);
\draw[-] (32) to (34);
\draw[dotted,-] (34) to (54);
\draw[-] (54) to (56);
\draw[-] (43) to (45);
\draw[dotted,-] (45) to (65);
\draw[-] (65) to (67);
\draw[dotted,-] (67) to (17);
\end{tikzpicture}

\begin{itemize}
\item Barry: ``No.''  
\item Abby: ``No.''  
\item Barry: ``No.''  
\item Abby: ``No.''
\end{itemize}

\begin{tikzpicture}[thick]
\node (17) at (0,0) {17};
\node (12) at (1,0) {12};
\node (54) at (4,0) {\color{white}54};
\node (45) at (-3,0) {\color{white}45};
\draw[-] (17) to (12);
\end{tikzpicture}

\medskip

Therefore, Abby was born on Monday!

\section{Dynamic Epistemic Logic}

Riddles about knowledge and ignorance such as those in this contribution have stood at the birth of the so-called {\em dynamic epistemic logic}, that is a modal logic of knowledge and change of knowledge. The models visualized are known as {\em Kripke models}. A well-known dynamic epistemic logic is {\em public announcement logic}. For example, in Cheryl's Birthday problem, Albert saying `I know that you don't know' can be formalized in the logic of knowledge as the formula $K_A \neg \bigvee_{p_d}  K_B p_d$, where $K_A$ stands for `Albert knows that' and $K_B$ for `Bernard knows that', $\neg$ is logical negation, $\bigvee$ stands for disjunction, and where $p_d$ is an atom ranging over the possible birthdays $d$. In the given model, this formula is true exactly on dates in July or August and false in dates in May and June. Albert's so-called public announcement of this formula therefore results in the model that is restricted to dates in July and August. Bernard subsequently saying `Now I know' corresponds to updating with (a public announcement of) the formula $\bigvee_{p_d}  K_B p_d$, and Albert's final announcement to an update with $\bigvee_{p_d}  K_A p_d$. For more information on dynamic epistemic logic see for example \cite{hvdetal.ajl:2005,moss.handbook:2015,baltagetal.stanford:2016}.
\bibliographystyle{eptcs}
\bibliography{biblio2017}
\end{document}